\documentclass[conference]{IEEEtran}
\IEEEoverridecommandlockouts

\usepackage{tikz}
\usepackage{float}
\usepackage{forest}
\usetikzlibrary{arrows.meta, shapes.geometric, positioning, fit, backgrounds}
\usepackage{booktabs}
\usepackage{array}
\usepackage{natbib}
\usepackage{amsmath,amssymb,amsfonts}
\usepackage{algorithmic}
\usepackage{graphicx}
\usepackage{textcomp}
\usepackage{caption}
\usepackage{xcolor}

\usepackage{hyperref}
\hypersetup{
    colorlinks=true,
    linkcolor=darkblue,
    citecolor=darkblue,
    urlcolor=darkblue
}
\definecolor{darkblue}{rgb}{0.0, 0.0, 0.55}

\def\BibTeX{{\rm B\kern-.05em{\sc i\kern-.025em b}\kern-.08em T\kern-.1667em\lower.7ex\hbox{E}\kern-.125emX}}
\begin{document}

\title{Achieving Trustworthy Real-Time Decision Support Systems with Low-Latency Interpretable AI Models
\thanks{\hspace*{-\parindent}\rule{3.8cm}{0.4pt} \\ 
$\ast$: Equal contribution. \\
$\dagger$: Corresponding author.}
}
\author{
\IEEEauthorblockN{Zechun Deng$^{\ast}$}
\IEEEauthorblockA{\textit{School of Physics and Astronomy} \\
\textit{University of Edinburgh}\\
Edinburgh, United Kingdom \\
z.deng-16@sms.ed.ac.uk}

\and

\IEEEauthorblockN{Ziwei Liu$^{\ast}$}
\IEEEauthorblockA{\textit{Siebel School of Computing and Data Science} \\
\textit{University of Illinois Urbana-Champaign}\\
Champaign, United States \\
ziweil2@illinois.edu}

\and

\IEEEauthorblockN{Ziqian Bi}
\IEEEauthorblockA{\textit{AI Agent Lab} \\
\textit{Vokram Group}\\
London, United Kingdom \\
bizi@vokram.com}

\and

\IEEEauthorblockN{Junhao Song$^{\dagger}$}
\IEEEauthorblockA{\textit{AI Agent Lab} \\
\textit{Vokram Group}\\
London, United Kingdom \\
ai-agent-lab@vokram.com}

\and

\IEEEauthorblockN{Chia Xin Liang}
\IEEEauthorblockA{\textit{AI Agent Lab} \\
\textit{Vokram Group}\\
London, United Kingdom \\
marcus.chia@ai-agent-lab.com}

\and

\IEEEauthorblockN{Joe Yeong}
\IEEEauthorblockA{\textit{Department of Anatomical Pathology} \\
\textit{Singapore General Hospital}\\
Singapore \\
yeongps@imcb.a-star.edu.sg}

\and

\IEEEauthorblockN{Xinyuan Song}
\IEEEauthorblockA{\textit{Department of Computer Science} \\
\textit{Emory University}\\
Atlanta, United States \\
xinyuan.song@emory.edu}

\and

\IEEEauthorblockN{Junfeng Hao}
\IEEEauthorblockA{\textit{AI Agent Lab} \\
\textit{Vokram Group}\\
London, United Kingdom \\
ygzhjf85@gmail.com}
}



\maketitle

\begin{abstract}
This paper investigates real-time decision support systems that leverage low-latency AI models, bringing together recent progress in holistic AI-driven decision tools, integration with Edge-IoT technologies, and approaches for effective human-AI teamwork. It looks into how large language models can assist decision-making, especially when resources are limited. The research also examines the effects of technical developments such as DeLLMa, methods for compressing models, and improvements for analytics on edge devices, while also addressing issues like limited resources and the need for adaptable frameworks. Through a detailed review, the paper offers practical perspectives on development strategies and areas of application, adding to the field by pointing out opportunities for more efficient and flexible AI-supported systems. The conclusions set the stage for future breakthroughs in this fast-changing area, highlighting how AI can reshape real-time decision support.
\end{abstract}

\begin{IEEEkeywords}
Interpretable AI, real-time decision support, human-AI teaming, edge computing, LLM fine-tuning, AI model compression, federated learning, decision making
\end{IEEEkeywords}

\section{Introduction and Motivation}

\begin{figure*}
  \includegraphics[width=\textwidth]{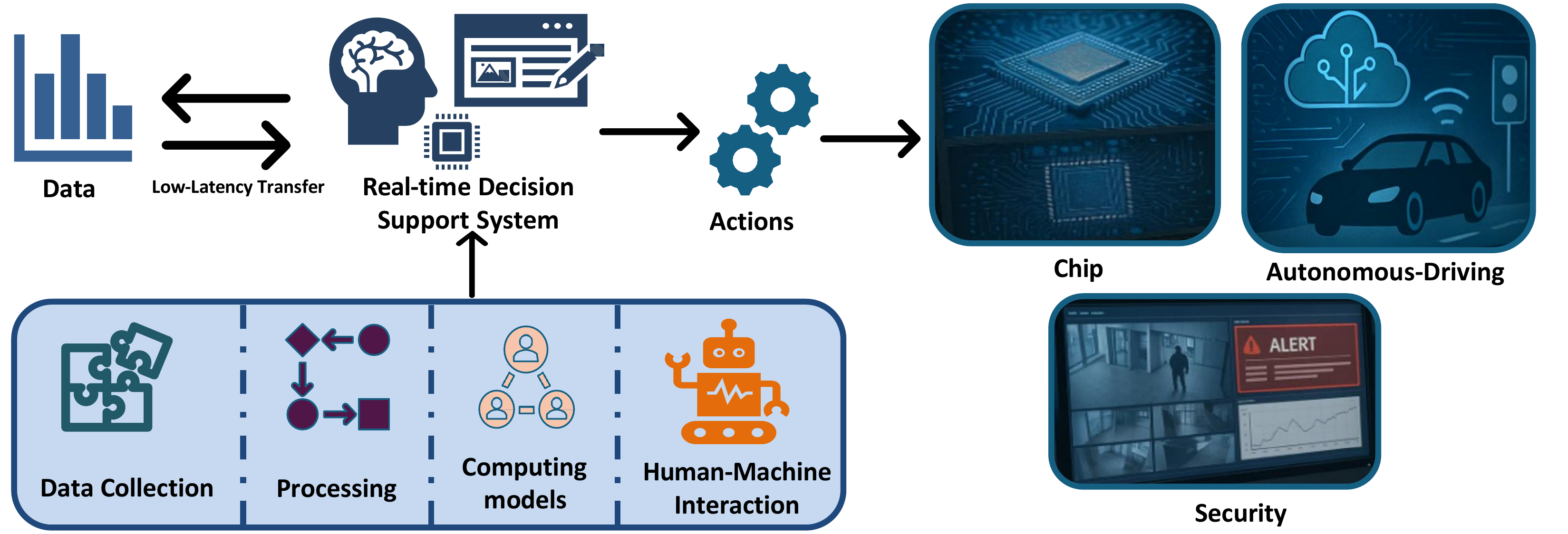}
  \captionof{figure}{\textbf{Examples of real-time decision support systems with low-latency explainable AI models for real-life applications.}}
  \label{teaser}
\end{figure*}

The field of artificial intelligence (AI) has experienced significant advances in recent years, transforming various domains such as security, healthcare, and general decision making \cite{1804.01396, 2410.23423}. Advancements in AI-enabling technologies have been crucial in the development of end-to-end AI systems, including data collection, processing, algorithms, computing models, and human-machine interaction methods \cite{1905.03592}. Automated operations using AI have also gained attention, with the potential to reduce human effort with better performance predictability and compatibility with the existing KPI framework \cite{2003.12808}. However, despite these advances, human-AI collaboration is still in its early stages, with a need for more deliberate consideration of interaction designs to achieve clear communication, trustworthiness, and collaboration \cite{2310.19778, 2401.05840}. The development of explainable AI models that can provide insights into their decision-making processes is also essential, enabling humans to trust and understand the outputs of AI systems \cite{2303.03511}.

The application of AI in decision support systems has shown great promise, for example, in the clinical diagnosis of COVID-19 \cite{2006.03434}. Nevertheless, the deployment of such systems is limited due to various challenges, including transparency of reasoning and explainability of outcome \cite{2004.13351}. The alignment between algorithmic outputs and human expectations is crucial, requiring a deep understanding of human intelligence and cognition. The need for real-time analytics and consideration of multiple aspects for multiple stakeholders in decision support systems has also been underscored by several literature on decision support systems in fisheries and aquaculture \cite{1611.08374}. These studies demonstrate that effective collaboration between humans and AI systems is crucial for improving decision-making outcomes, particularly in high-stakes settings \cite{2210.12849}. Indeed, this review emphasizes the importance of human-AI collaboration, such as the use of behavior descriptions to increase human-AI accuracy \cite{2301.06937} and the development of a unified framework to predict human behavior in AI-assisted decision making \cite{2401.05840}. 

The availability and quality of data is another area calling for attention. Over the recent years, pretrained Large Language Models (LLMs) have been increasingly recognized for their potential in supporting decision-making processes, particularly in low-resource contexts where data is scarce or uncertain \cite{2402.02392}. Nonetheless, determinants such as technological literacy, psychological factors, and domain-specific aspects can impact the decision-making processes with LLMs, similar to other forms of technology \cite{2402.17385}. The same studies have also highlighted the importance of fine-tuning LLMs to adapt them to specific tasks and domains, including decision-making under uncertainty. 

Edge Computing and IoT Integration for Real-Time Analytics serves as a key aspect of Real-Time Decision Support Systems, enabling organizations to make informed decisions by analyzing data as they are generated and processed. The use of edge computing and IoT devices enables efficient and scalable AI deployments, with applications in areas such as visual quality inspection \cite{2501.17062}, air quality prediction, and scientific experimentation \cite{2207.07958}. Moreover, various optimization techniques, such as federated learning, cross-level optimization and model-adaptive system scheduling, have been employed to achieve resource-efficient AIoT systems \cite{2204.10183}. The use of machine learning models, including deep neural networks, has become ubiquitous in AIoT systems, enabling predictive analytics and decision making \cite{2405.12236}.

In conclusion, the motivation of this work is spurred by the significant advancements of AI in recent years, with applications in various domains. However, challenges and opportunities must be openly discussed in areas such as human-AI collaboration models, algorithmic solutions, and hardware support. The findings of this study will serve as implications for the development of effective real-time decision support systems using low-latency AI models. By addressing the limitations and challenges identified in these papers, researchers can develop more effective and generalizable approaches to supporting human decision-making in complex, dynamic environments, ultimately leading to improved outcomes and decision-making capabilities.

\section{Human-AI Cooperation in Decision-Making: Current State and Challenges}

\renewcommand{\arraystretch}{1.5}
\begin{table*}[ht]
\centering
\caption{\textbf{Architecture Comparison of mainstream Generative AI models}}
\begin{tabular}{|>{\raggedright\arraybackslash}p{2.5cm} 
                |>{\raggedright\arraybackslash}p{3cm} 
                |>{\raggedright\arraybackslash}p{2.8cm} 
                |>{\raggedright\arraybackslash}p{2.5cm} 
                |>{\raggedright\arraybackslash}p{5cm}|}
\hline
\textbf{Model} & \textbf{Architecture Type} & \textbf{Parameters (B)} & \textbf{Context Length} & \textbf{Notable Features} \\
\hline
\raisebox{-0.7
em}{\includegraphics[height=0.35cm]{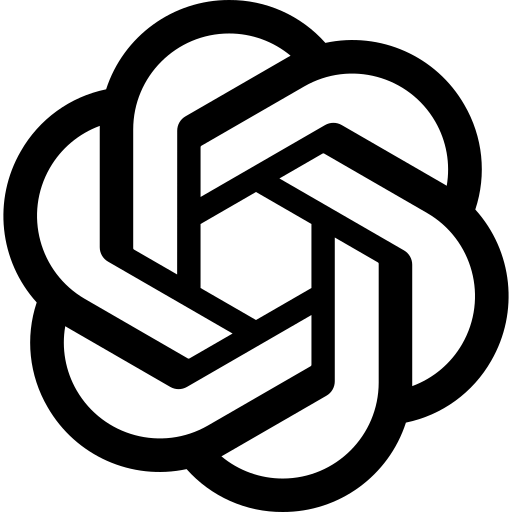}} \vspace{2pt} \raisebox{-0.5em} {ChatGPT 4.5} & Transformer (Mixture of Experts, likely) & Undisclosed (estimated 1.5 T+) & 128K & Efficient inference, multi-modal support, fine-tuned for chat, visual storytelling \\
\hline
\raisebox{-0.75em}{\includegraphics[height=0.35cm]{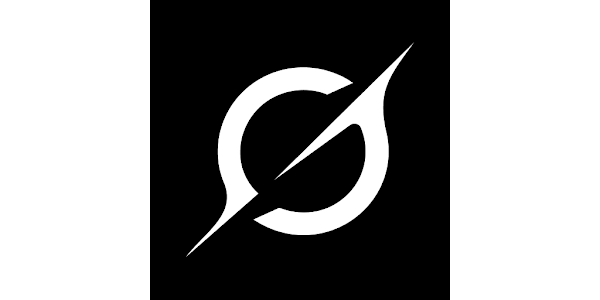}} \vspace{2pt} \raisebox{-0.5em} {Grok-3} & Transformer-based & Not officially disclosed & Estimated 1M & Designed for intelligent, conversational, and problem-solving capabilities \\
\hline
\raisebox{-0.75em}{\includegraphics[height=0.35cm]{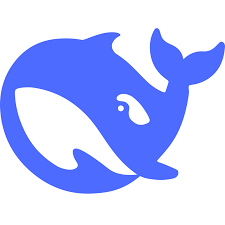}} \vspace{2pt} \raisebox{-0.5em} {DeepSeek-V3} & Transformer, dense + sparse Mixture of Experts & 236B (combined) & 128k & Fast inference speed, Memory-efficient design, strong math/coding capabilities \\
\hline
\raisebox{-0.7em}{\includegraphics[height=0.35cm]{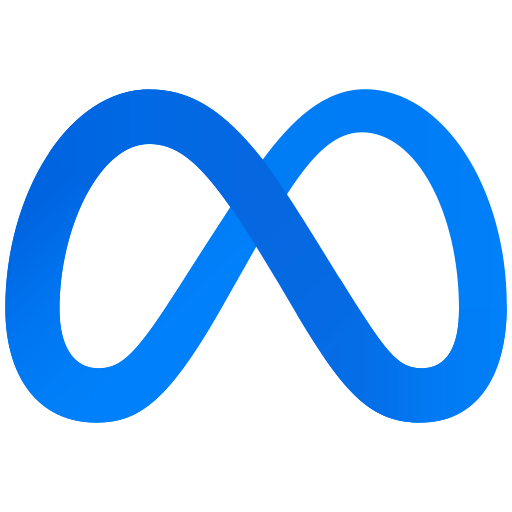}} \vspace{2pt} \raisebox{-0.5em} {LLaMA 4} & Transformer, Decoder-only & Estimated 400B (largest variant) & 10M & Open-weight (Meta), efficient scaling, strong multilingual support \\
\hline
\raisebox{-0.75em}{\includegraphics[height=0.35cm]{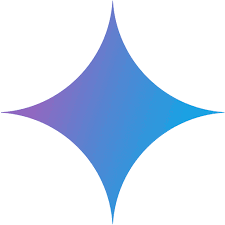}} \vspace{2pt} \raisebox{-0.5em} {Gemini 2.5} & Transformer, Mixture-of-Experts & Not disclosed & 1M & Reasoning through their thoughts before responding \\
\hline
\end{tabular}
\end{table*}

The topic of human-AI cooperation in decision making has garnered significant attention in recent years, with a growing body of research underscoring its importance \cite{2012.06034}. Recent studies have explored various aspects of human-AI cooperation, including the impact of user expertise and algorithmic tuning on joint decision making \cite{2208.07960}, the importance of decoding AI's judgment in predicting human behavior \cite{2401.05840}, and the value of information sharing in human-AI decision making \cite{2502.06152}. Additionally, researchers have investigated the requirements of clinicians for explainable AI in healthcare decision making \cite{2411.11774}. These developments underscore the growing recognition of the importance of human-AI cooperation in decision making and highlight the need for continued research in this area.

However, the current state of human-AI collaboration is characterized by simplistic interaction paradigms, which limit the potential for effective collaboration \cite{2310.19778}. This limitation is further compounded by the need for research in "AI and Cooperation" to understand how systems of AIs and humans can engender cooperative behavior \cite{2012.06034}. Given this, a taxonomy of interaction patterns in AI-assisted decision making has been introduced, highlighting the need for more deliberate consideration of interaction designs to achieve clear communication, trustworthiness, and collaboration. For example, recent research has made significant contributions to exploring both learning-to-defer (L2D) and learning-to-complement (L2C) approaches \cite{2411.11976}. The Coverage-constrained Learning to Defer and Complement with Specific Experts (CL2DC) method effectively explores these approaches under diverse expert knowledge, achieving superior performance compared to state-of-the-art methods \cite{2411.11976}. There are also debates around centralized vs. decentralized decision-making \cite{2104.14089}, and prescriptive vs. adaptive methods \cite{2302.02944}. Furthermore, researchers have proposed both human-centric \cite{2403.05911} and task-centric objectives \cite{2310.19778}, highlighting the complexity of balancing these priorities in human-AI cooperation. 

For example, a model of human reliance behavior can be represented using a supervised fine-tuning loss such as:

\begin{equation}
\mathcal{L}_{\text{SFT}}(\theta) = - \sum_{i=1}^{N} \log p_\theta(y_i \mid x_i),
\label{eq1}
\end{equation}
which captures how likely the AI’s response \( y_i \) aligns with human-annotated outputs given an instruction \( x_i \). Furthermore, when training the AI to optimize for human preferences, reinforcement learning from human feedback (RLHF) is often applied using the loss:

\begin{equation}
\mathcal{L}_{\text{RL}}(\theta) = - \mathbb{E}_{y \sim \pi_\theta} [r(y)],
\label{eq2}
\end{equation}
where the model’s policy \( \pi_\theta\) is adjusted to maximize reward signals derived from human feedback. To ensure stability and prevent drift from the original model, KL-regularized objectives like:
\begin{equation}
\mathcal{L}_{\text{PPO}} = - \mathbb{E}_{y \sim \pi_\theta} \left[ \frac{\pi_\theta(y)}{\pi_{\text{ref}}(y)} \hat{A}(y) - \beta \, \text{KL}[\pi_\theta \parallel \pi_{\text{ref}}] \right].
\label{eq3}
\end{equation}
And when learning from human preferences directly, a pairwise preference modeling loss can be applied:

\begin{equation}
    \mathcal{L}_{\text{PM}}(\phi) = - \sum_{(y^+, y^-) \in D} \log \sigma(r_\phi(y^+) - r_\phi(y^-)),
\label{eq4}
\end{equation}
which teaches the AI to rank preferred responses higher and higher. Furthermore, when selecting cues to train instruction following models, metrics based on instruction similarity
\begin{equation}
    s(x) = \frac{1}{k} \sum_{i=1}^{k} \text{sim}(x, x_i).
\label{eq5}
\end{equation}
which can ensure the models are diverse and informative. Together, these models form the computational backbone that captures how humans rely on AI. This process is influenced not only by model accuracy but also by trust, feedback, and contextual interpretation. Such neural models of human behavior can take task characteristics, AI-generated suggestions, and interpretability cues as input and predict the user's level of reliance based on dynamically learned trust states and experience-based adjustments. This example and the corresponding equations \ref{eq1}, \ref{eq2}, \ref{eq3}, \ref{eq4}, \ref{eq5} are mentioned in the article by \cite{2023.25748}.


\begin{figure*}[ht]
    \centering
    \includegraphics[width=0.88\linewidth]{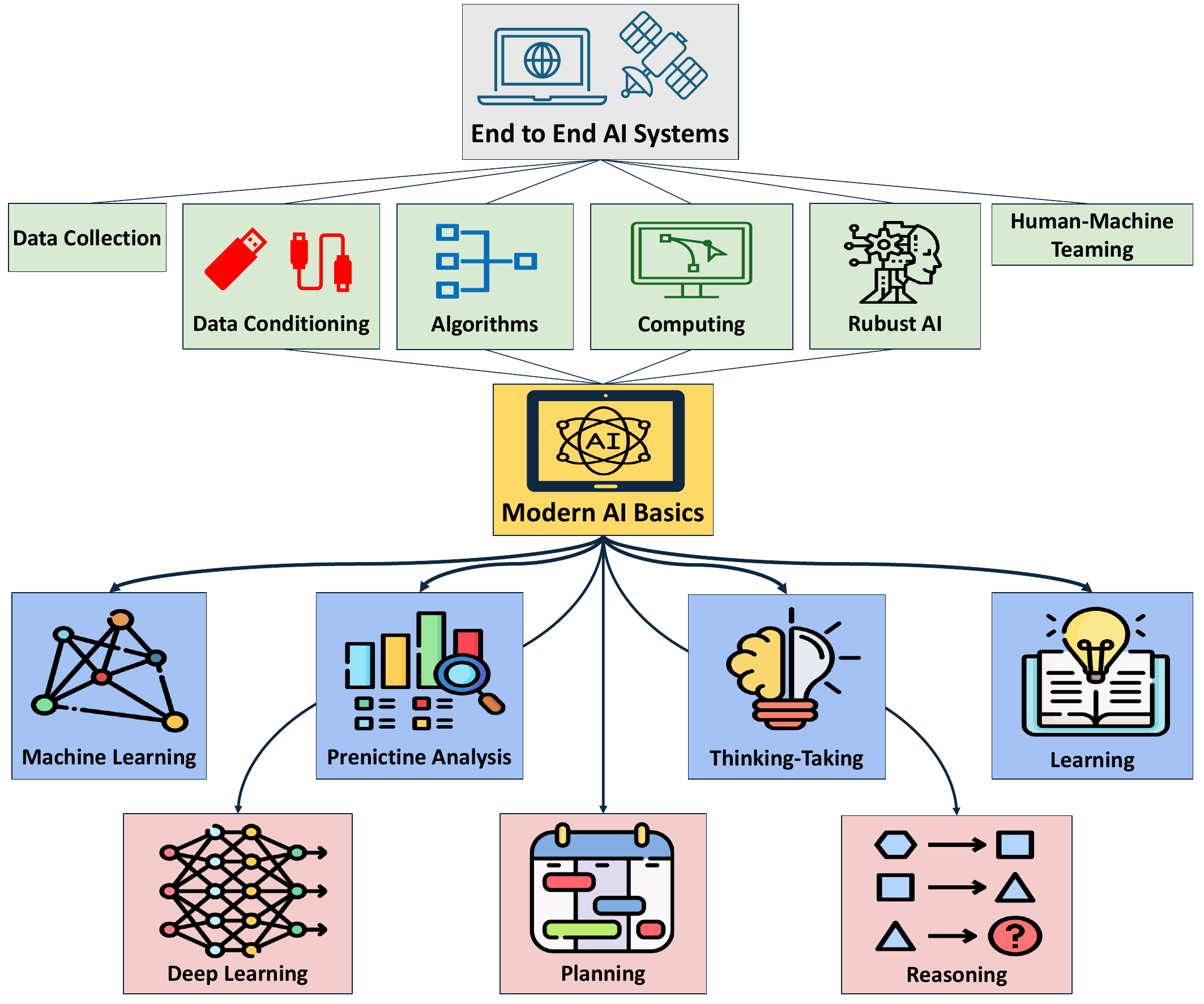}
    \caption{\textbf{Main components of an End-to-End System and functions of modern AI models.}}
    \label{fig:e2e}
\end{figure*}

The success of AI-driven decision making also relies heavily on the ability of human to make informed decisions in uncertain environments, where data quality and annotation accuracy can be compromised. Recent studies have made significant contributions to this area, providing valuable insights into the challenges and opportunities associated with decision-making under uncertainty. For instance, the importance of decision-theoretic reasoning in uncertain environments is highlighted in \cite{1304.3442}, which demonstrates the effectiveness of exact reasoning under uncertainty. Dynamic information selection has also been introduced as a novel framework for AI assistance, with methods such as dynamic Information Sub-Selection (DISS) discussed by \cite{2410.23423} tailoring information processing on a per-instance basis to generate superior performance. Multi-agent approaches have also been shown to effectively represent complex information, with a multi-agent system composed of factual agents developing through interactions and comparisons of semantic features \cite{2104.14089}. Development of more flexible AI systems that can incorporate both 'fast' and 'slow thinking', utilizing previous experience and deliberate reasoning, has been identified as a potential solution \cite{2201.07050} in specialized environments such as driving, which is particularly relevant to real-time decision support systems where adaptability is crucial. 

Last but not least, the role of trust, transparency, and accountability in human-AI decision making is also a critical consideration, as emphasized in \cite{2310.19778} and \cite{2012.06034}. Explainability and interpretability are the foundation of building AI models that can give trustworthy insights into the decision-making process, especially as studies \cite{2411.10461} have demonstrated that human behavior can be influenced by manipulated explanations in AI-assisted contexts, which can emphasize the need for transparent and interpretable AI systems. The importance of this aspect is further emphasized by \cite{2003.07370}, which proposes developing interpretable methods to explain machine predictions, essential for improving human decision-making with AI assistance in real-time scenarios. Additionally, \cite{2112.11471} highlights the need for empirical approaches to form a foundational understanding of how humans interact with AI to make decisions. Studies such as \cite{2301.06937} has also demonstrated that providing people with descriptions of AI behavior can increase human-AI accuracy by helping them identify AI failures and increasing reliance on the AI when it is more accurate. Some studies employed a contrasting viewpoint, however. For instance, \cite{2201.07050} focuses on developing "black-box" models that predict human behavior. In the context of reliable decision support systems, there is a need for both types of model, as well as hybrid approaches that balance prediction accuracy and interpretability.

In conclusion, the domain of human-AI cooperation in decision-making is characterized by significant challenges and opportunities. While simplistic interaction paradigms currently limit the potential for effective collaboration, research in "AI and Cooperation" and empirical studies on human-AI decision making can help overcome these limitations. By developing more collaborative interaction paradigms, common frameworks for empirical studies, and effective methods for optimizing AI' capacity and explainability, researchers can unlock the full potential of human-AI cooperation and create more effective decision-making systems.

\section{Large Language Models and their Applications in Low-Resource Contexts}

Recent studies have proposed frameworks to enhance decision-making accuracy using LLMs, such as DeLLMa, which achieves up to a 40\% increase in accuracy over competing methods by integrating principles from decision theory and utility theory \cite{2402.02392}. Furthermore, research has explored the development of efficient LLMs, with a focus on reducing model size and computational resources while maintaining performance \cite{2309.06589}. For example, the study ``Do Generative Large Language Models need billions of parameters?" presents novel systems and methodologies for developing efficient LLMs, highlighting trade-offs between model size, performance, and computational resources \cite{2309.06589}.

\begin{figure*}[ht]
    \centering
    \includegraphics[width=0.9\linewidth]{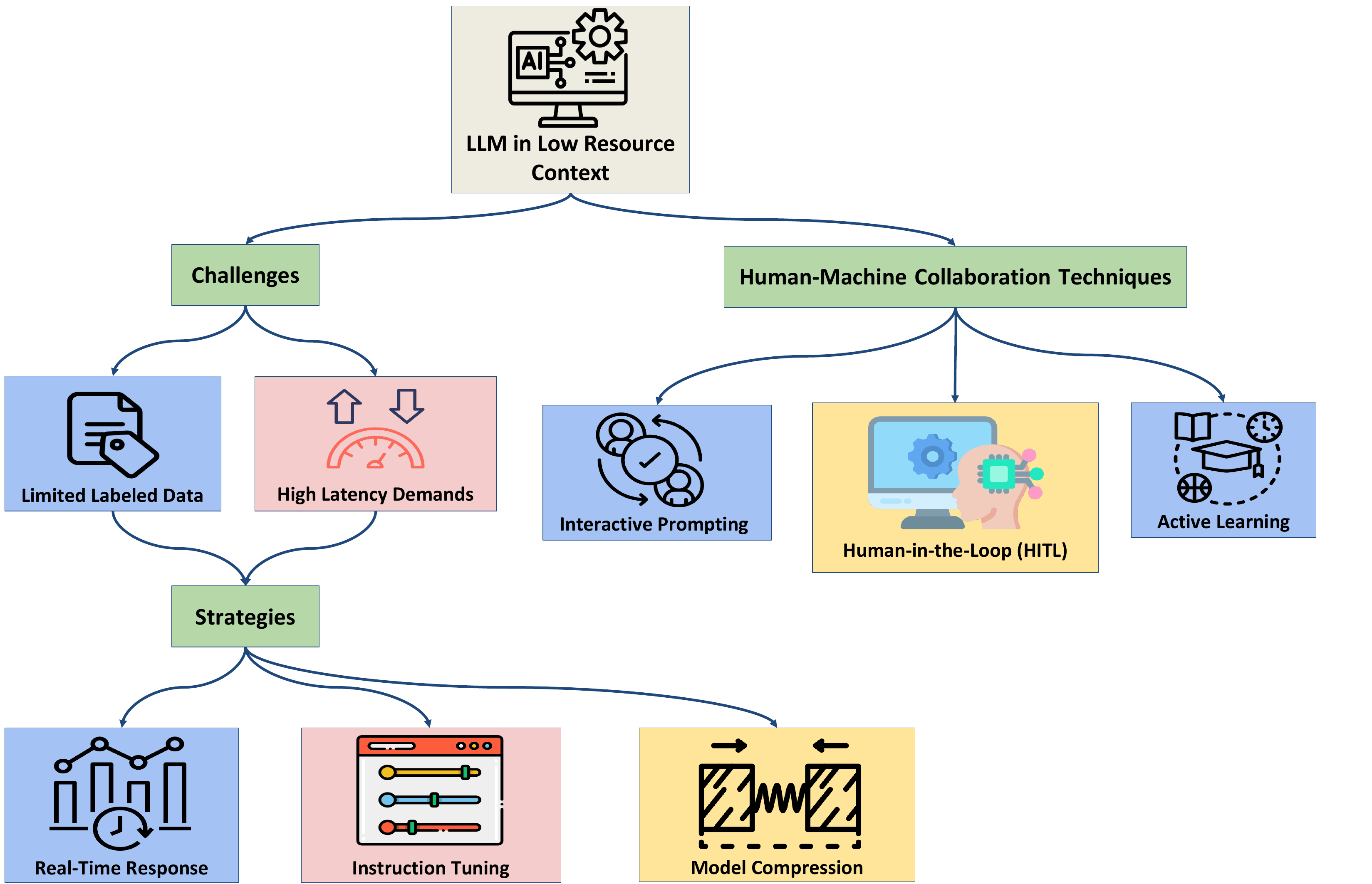}
    \caption{\textbf{Main challenges with strategies and techniques when using Large Language Models in Low-Resource Contexts.}}
    \label{fig:llm}
\end{figure*}

The application of LLMs in low-resource contexts is not limited to decision-making support; they have also been explored for their potential in language adaptation and benchmarking \cite{2405.04685}, as well as in multilingual settings \cite{2410.17532}. In the context of language education, foundation models have been proposed as a means to support low-resource language learning \cite{2412.04774}. Moreover, LLMs have been fine-tuned for specific domains, such as robotic decision-making \cite{2501.16899} and legal understanding \cite{2412.14771}, demonstrating their versatility and potential for adaptation.

\begin{figure*}[ht]
    \centering
    \includegraphics[width=0.85\linewidth]{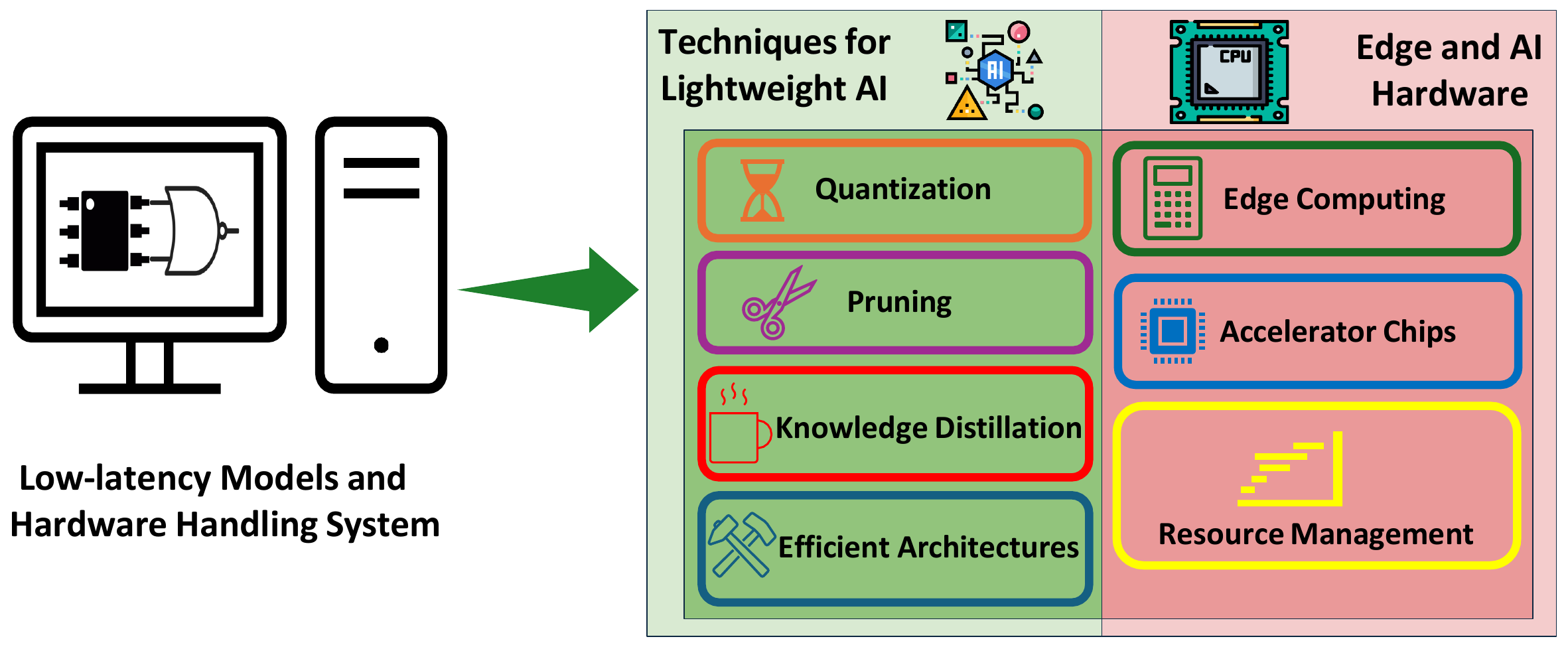}
    \caption{\textbf{Technical and hardware foundations of Low-latency Models and Decision Handling System.}}
    \label{fig:eh}
\end{figure*}

A common theme across these studies is the emphasis on improving the efficiency and effectiveness of LLMs, whether through novel methodologies, frameworks, or understanding determinants \cite{2402.17385}. The importance of transparency, interpretability, and accountability in LLM-assisted decision-making is also highlighted, with research emphasizing the need for a deeper understanding of LLMs' limitations and challenges \cite{2023.47711}. For example, the study "Determinants of LLM-assisted Decision-Making" provides a comprehensive literature analysis on the influencing factors of LLM-assisted decision-making, identifying technological, psychological, and decision-specific determinants \cite{2402.17385}.

The technical details and methodologies employed in these studies vary, with some utilizing multi-step reasoning procedures \cite{2402.02392}, while others explore novel methods for reducing model size \cite{2309.06589} or develop dependency frameworks to systematize interactions between determinants \cite{2402.17385}. The use of variability models to depict factors guiding model selection recommendations has also been explored \cite{2311.11516}. These technical advancements have significant implications for the development of real-time decision support systems, particularly in low-latency applications where efficient and effective decision-making is critical.

Despite these developments, limitations and challenges remain, including the potential for poor performance in complex decision-making tasks \cite{2402.02392} and the need for more efficient models \cite{2309.06589}. The need for more efficient and effective methods for incorporating LLMs into decision making processes is highlighted in papers such as \cite{2404.11973} and \cite{2311.06622}. The complexity of interactions between various determinants impacting decision-making with LLM support is also a significant challenge \cite{2402.17385}. Furthermore, the importance of considering ethical considerations and assumptions about data in the model selection process has been highlighted \cite{2311.11516}.

In conclusion, recent research has demonstrated the potential of LLMs in supporting decision-making processes in low-resource contexts, with a focus on improving efficiency, effectiveness, and transparency. The development of efficient LLMs, fine-tuned for specific domains, and the exploration of novel methodologies and frameworks have significant implications for real-time decision support systems. However, limitations and challenges remain, emphasizing the need for continued research and development in this area to realize the potential of LLMs in low-resource contexts fully \cite{2023.47711}. As the field continues to evolve, it is essential to consider the connections between LLMs, decision-making, and real-time applications, ultimately leading to more informed and effective decision-making processes \cite{2402.02392}.

\section{Practical Considerations in Fine-Tuning LLMs for Domain-Specific Applications}

Fine-tuning Large Language Models (LLMs) for domain-specific applications has emerged as a crucial aspect of developing real-time decision support systems using low-latency AI models \cite{2408.13296}. The importance of optimizing LLMs for real-world applications is further emphasized by papers such as \cite{2406.12125} and \cite{2407.06486}, which propose approaches that leverage online model selection algorithms and simulations to efficiently incorporate LLMs into sequential decision making. Techniques such as Low-Rank Adaptation (LoRA) and Half Fine-Tuning have also been explored to address this challenge \cite{2410.20297}, \cite{2404.14779}.

Applications of fine-tuned LLMs span across multiple industries. Papers such as \cite{2401.02981} and \cite{2411.03350} emphasize the role of fine-tuned LLMs in domains including medical applications \cite{2404.14779} and scientific knowledge extraction \cite{2408.04651}. Despite the advancements in fine-tuning LLMs for domain-specific applications, significant challenges and limitations remain. Fine-tuning LLMs can be computationally expensive and require substantial amounts of labeled data \cite{2409.03563}. Moreover, current LLMs may not be suitable for certain types of decision making tasks, particularly those requiring a nuanced understanding of various outcomes and consequences even with fine-tuning \cite{2408.06087}.

To sum up, the analysis of relevant papers highlights the importance of fine-tuning LLMs for domain-specific applications. The proposed frameworks and approaches, such as DeLLMa and online model selection algorithms, demonstrate the potential of fine-tuned LLMs in improving decision-making accuracy \cite{2402.02392}, \cite{2406.12125}. Future research directions should focus on developing more efficient and effective methods for fine-tuning LLMs, as well as exploring new applications and domains where these models can be leveraged to enable smart decision-making \cite{2411.17722}, \cite{2408.04651}. The challenges and limitations associated with fine-tuning LLMs, including computational efficiency, transparency, and explainability, must be addressed to fully realize the benefits of these models in real-time decision support systems \cite{2408.13296}. 



\section{Edge Computing and IoT Integration for Real-Time Analytics}

\begin{figure*}[ht]
    \centering
    \includegraphics[width=0.8\linewidth]{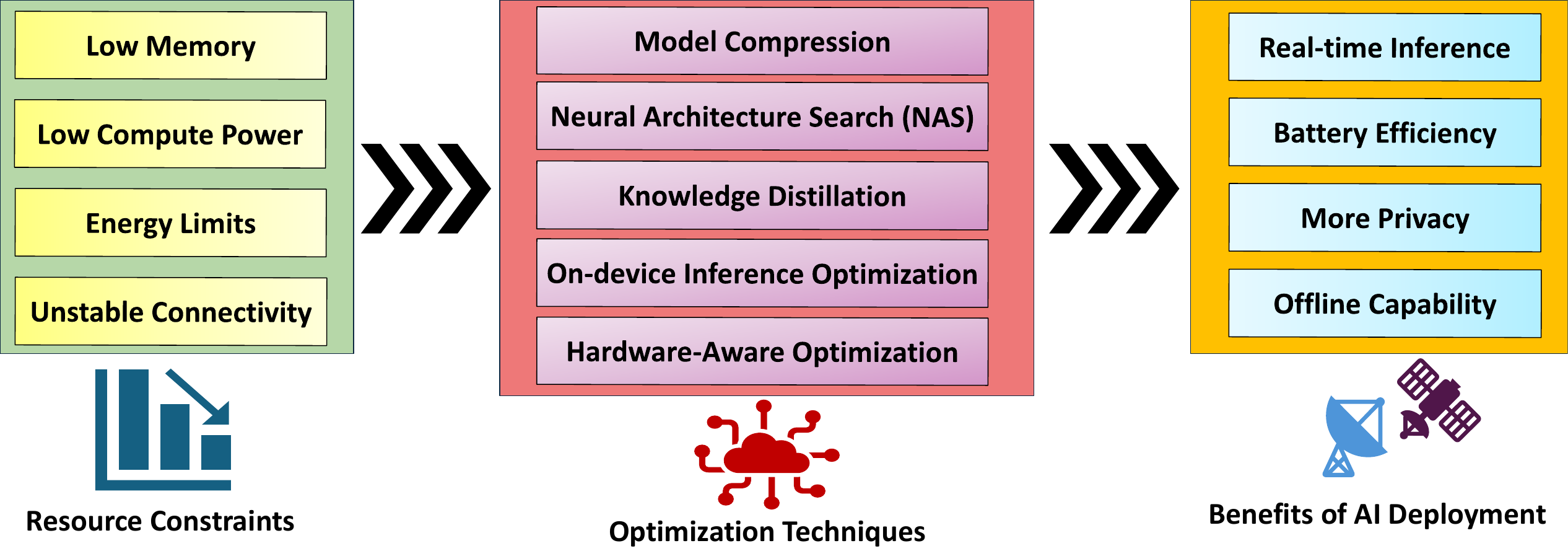}
    \caption{\textbf{How optimization techniques help overcome resource limitations of IoT devices when deploying AI models.}}
    \label{fig:embedded_system}
\end{figure*}

The integration of Edge Computing and IoT allows for real-time decision making using artificial intelligence, as seen in \cite{2105.01798} and \cite{2408.30574}. Edge computing enables efficient operations by processing data closer to the source, thereby minimizing communication overheads \cite{2405.12236}. The paper ``Reliable Fleet Analytics for Edge IoT Solutions" proposes a framework for facilitating machine learning at the edge, enabling continuous delivery, deployment, and monitoring of machine learning models \cite{2101.04414}. Furthermore, the integration of LLMs with IoT has been explored, highlighting opportunities for advanced decision-making and contextual understanding \cite{2411.17722}. These approaches highlight the importance of low-latency communication and real-time decision making in IoT devices with edge computing, a common theme across all papers.

Edge computing that focuses on moving computing closer to the data source is a critical solution to reduce latency and improve real-time decision making \cite{2105.01798}. Recent studies have proposed various frameworks and techniques for deploying and managing machine learning models at the edge, including EdgeMLOps \cite{2501.17062}, FastML Science Benchmarks \cite{2207.07958}. On the other hand, model optimization, deployment, and lifecycle management are complex in edge environments, where resource constraints, such as limited computational power and memory, pose significant challenges \cite{2107.06835, 2302.08571, 2408.30574}. Different papers propose varying approaches to deploying machine learning models at the edge, including partitioning, quantization, and early exit \cite{2411.00907}. Some studies focus on the use of cloud deployments, while others emphasize the importance of edge-tier deployments with higher network and computational capabilities \cite{2212.03332}. The choice of deployment strategy depends on factors such as model complexity, input data size, and available resources \cite{2009.00803}.

Additionally, executing advanced and sophisticated analytic algorithms on edge devices is difficult, and there is a need for efficient and effective query allocation and data analysis on edge computing nodes \cite{2008.05427}. A review of edge analytics, provided in \cite{2107.06835}, highlights the issues, challenges, opportunities, and future directions of edge analytics, including its applications in various areas such as retail, agriculture, industry, and healthcare. Software-defined edge computing, as introduced in \cite{2104.11645}, provides a promising architecture to support IoT data analysis, allowing for more efficient and flexible data processing. Furthermore, an intelligent edge-centric queries allocation scheme based on ensemble models, proposed in \cite{2008.05427}, optimizes query allocation in edge computing nodes, addressing the need for efficient and effective data analysis. Finally, other papers such as \cite{2403.09141} and \cite{2103.06518} discuss the uncertainty estimation in multi-agent distributed learning and data collection and utilization frameworks for edge AI applications. 

In conclusion, the integration of Edge Computing and IoT for real-time analytics has the potential to revolutionize various industries by enabling real-time decision making using artificial intelligence. The implications of these developments are significant, with potential applications in various domains such as smart cities, healthcare, and industrial automation \cite{1903.10583}. Meanwhile, limitations and challenges persist, including resource constraints, communication overheads, scalability, reliability, security, and privacy concerns \cite{2011.08612}. Addressing these challenges will require continued research and innovation in AI-driven IoT solutions. 

\section{Optimization Techniques for Efficient AIoT Deployment}

The integration of Artificial Intelligence (AI) with Internet of Things (IoT) has given rise to numerous opportunities for real-time decision-making, improved efficiency, and enhanced automation \cite{2011.08612}. However, this convergence known as AIoT poses significant challenges and research gaps that need to be addressed \cite{2411.17722}. One of the primary concerns is the resource constraint of IoT devices, which limits the deployment of complex AI models \cite{2204.10183}. To mitigate this issue, researchers have proposed various optimization techniques such as model compression, pruning, and knowledge distillation \cite{2204.10183}. For instance, the paper "Multi-Component Optimization and Efficient Deployment of Neural-Networks on Resource-Constrained IoT Hardware" presents an end-to-end optimization sequence that achieves significant compression, accuracy improvement, and inference speedup \cite{2204.10183}. The paper "Pervasive AI for IoT applications: A Survey on Resource-efficient Distributed Artificial Intelligence" provides a comprehensive survey of techniques for overcoming resource challenges in pervasive AI systems \cite{2105.01798}. Additionally, the need for distributed algorithms and communication-efficient techniques is a recurring theme, reflecting the complexity of IoT environments \cite{2404.04205}. 

Recent research has highlighted the importance of federated learning in enabling efficient and robust operations \cite{2101.04414}. The proposed framework facilitates continuous delivery, deployment, and monitoring of machine learning models, which is essential for real-time decision making in AIoT systems. Different approaches to federated learning have been proposed, including Split Federated Learning (SFL) and Sliding Split Federated Learning (S$^2$FL), each with varying degrees of emphasis on accuracy and efficiency \cite{2310.07497}. By adopting an adaptive sliding model split strategy and a data balance-based training mechanism, these techniques has led to significant inference accuracy improvement and training acceleration \cite{2311.13163}. Cross-level optimization is also essential in achieving this goal, involving the joint optimization of resource-friendly deep learning (DL) models and model-adaptive system scheduling \cite{2305.10437}. Additionally, formal specifications, such as mathematical formalisms, can be used to deploy effective IoT architectures, providing a rigorous framework for describing and optimizing AIoT systems \cite{1907.09059}. 

Deep Reinforcement Learning (DRL) is another promising method for achieving greater autonomy in AIoT, with applications in various domains, including autonomous Internet of Things \cite{1908.01656}. However, the deployment of DRL models in resource-constrained IoT devices requires innovative solutions to address issues such as scalability, security, and efficiency \cite{2207.08226}. Distributed computing and edge computing are the key factors in accommodating the large number of IoT devices and vast amounts of data transfer \cite{2407.10987}. The emergence of new architectures, such as fog computing, has shown promises in computing infrastructure closer to data sources, enabling real-time processing and decision making \cite{2210.01985}. The paper "Fully Distributed Fog Load Balancing with Multi-Agent Reinforcement Learning" proposes a fully distributed approach to fog load balancing using multi-agent reinforcement learning \cite{2405.12236}. Lastly, other emerging technologies like 6G and blockchain will also be crucial in shaping the future of AIoT applications \cite{2501.12420}.

In conclusion, efficient AIoT operations are a vital component of real-time decision support systems using low-latency AI Models. Recent research has highlighted the importance of edge computing, federated learning, and cross-level optimization in enabling efficient and robust operations. However, several challenges remain, including scalability, resource constraints, security, and autonomy, which must be addressed to realize the full potential of AIoT systems \cite{2110.01234}. Ensuring secure communication and data protection in AIoT systems is also critical, particularly in applications involving sensitive information \cite{2411.17722}. As research continues to evolve, we can expect significant advancements in the development of efficient and autonomous AIoT systems, enabling real-time decision making and transforming various aspects of people's lives.

\begin{figure*}
  \centering
\begin{forest}
  for tree={
    grow=east,
    reversed=true,
    anchor=base west,
    parent anchor=east,
    child anchor=west,
    base=left,
    rectangle,
    draw=black,
    rounded corners,
    align=left,
    minimum width=3em,
    edge+={gray, line width=1pt},
    inner xsep=4pt,
    inner ysep=1pt,
    font=\small,
  },
  where level=1{fill=green!20}{},
  where level=2{fill=cyan!10}{},
  where level=3{yshift=0.3pt, fill=purple!10}{},
  where level=4{yshift=0.3pt, fill=orange!20}{},
  where level=5{yshift=0.3pt, fill=gray!10}{},
  [\rotatebox{90}{\textbf{Real-Time Decision Support Systems}},
    [AI Advancements,
        [End-to-End AI,
            [Component Types,
                [Data Collection \\
                 Data Conditioning \\
                 Algorithms \\
                 Computing \\
                 Robust AI
                ]
            ]
            [Cognitive Models,
                [Fast-Slow Thinking \\
                 Metacognition
                ]
            ]
        ]
        [Human-AI Collaboration,
            [Learning Approaches,
                [Learning-to-Defer \\
                 Learning-to-Complement \\
                 CL2DC
                ]
            ]
            [Trust Building,
                [Transparency \\
                 Explainability \\
                 Behavior Descriptions
                ]
            ]
        ]
        [Decision Frameworks,
            [Optimization Methods,
                [Offline Reinforcement Learning \\
                 DeLLMa Framework \\
                 Dynamic Information Sub-Selection
                ]
            ]
            [Domain Applications,
                [Clinical Decision Support \\
                 COVID-19 Systems
                ]
            ]
        ]
        [Human-Centered,
            [Design Approaches,
                [Interdisciplinary Methods \\
                 Alignment with Human Expectations
                ]
            ]
            [Coverage-Constrained,
                [Coverage-Constrained Cooperation
                ]
            ]
        ]
    ]
    [Edge Computing,
        [Edge Analytics,
            [Application Areas,
                [Retail \\
                 Agriculture \\
                 Industry \\
                 Healthcare
                ]
            ]
            [Framework Types,
                [Software-defined Edge Computing \\
                 Fleet Analytics Framework
                ]
            ]
        ]
        [Query Allocation,
            [Optimization Models,
                [Ensemble Models \\
                 Intelligent Edge-Centric Scheme
                ]
            ]
        ]
        [IoT Integration,
            [Real-Time Monitoring,
                [Deep Reinforcement Learning \\
                 Edge-Cloud Collaboration
                ]
            ]
            [Deployment Models,
                [Lightweight Deep Learning \\
                 Edge-Tier Deployment
                ]
            ]
            [Resource Optimization,
                [Quantization Methods \\
                 Model Compression \\
                 Cross-level Optimization
                ]
            ]
        ]
    ]
  ]
\end{forest}
\caption{\textbf{Taxonomy of Real-Time Decision Support Systems Using Low-Latency AI Models (Part 1).}}
\end{figure*}
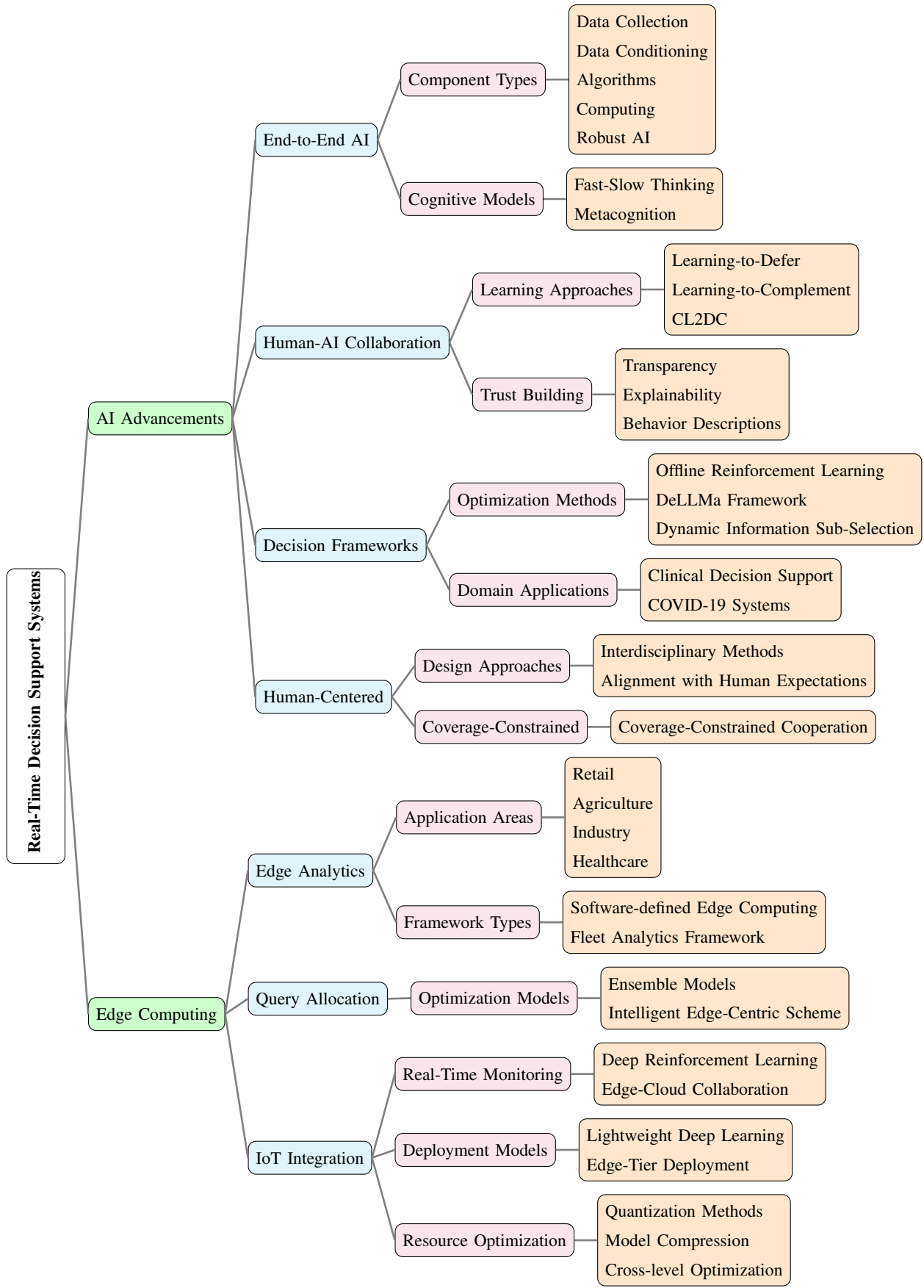

\begin{figure*}
  \centering
\begin{forest}
  for tree={
    grow=east,
    reversed=true,
    anchor=base west,
    parent anchor=east,
    child anchor=west,
    base=left,
    rectangle,
    draw=black,
    rounded corners,
    align=left,
    minimum width=3em,
    edge+={gray, line width=1pt},
    inner xsep=4pt,
    inner ysep=1pt,
    font=\small,
  },
  where level=1{fill=green!20}{},
  where level=2{fill=cyan!10}{},
  where level=3{yshift=0.3pt, fill=purple!10}{},
  where level=4{yshift=0.3pt, fill=orange!20}{},
  where level=5{yshift=0.3pt, fill=gray!10}{},
  [\rotatebox{90}{\textbf{Real-Time Decision Support Systems}},
    [AI Models,
        [Large Language Models,
            [Fine-Tuning Techniques,
                [Low-Rank Adaptation (LoRA) \\
                 Half Fine-Tuning \\
                 Domain-Specific Adaptation
                ]
            ]
            [Low-Resource Contexts,
                [Language Adaptation \\
                 Multilingual Models \\
                 Foundation Models
                ]
            ]
            [Decision Support,
                [Medical LLMs \\
                 Legal Understanding \\
                 Robotic Decision-Making
                ]
            ]
        ]
        [Federated Learning,
            [Algorithm Types,
                [Split Federated Learning (SFL) \\
                 Sliding Split Federated Learning (S²FL)
                ]
            ]
            [Data Management,
                [Data Balance-based Training \\
                 Adaptive Sliding Model Split
                ]
            ]
        ]
        [Edge MLOps,
            [Model Optimization,
                [Static signed-int8 \\
                 Dynamic signed-int8 \\
                 Knowledge Distillation
                ]
            ]
            [Deployment Strategies,
                [Partitioning \\
                 Early Exit \\
                 Pruning
                ]
            ]
        ]
    ]
  ]
\end{forest}
\caption{\textbf{Taxonomy of Real-Time Decision Support Systems Using Low-Latency AI Models (Part 2).}}
\end{figure*}
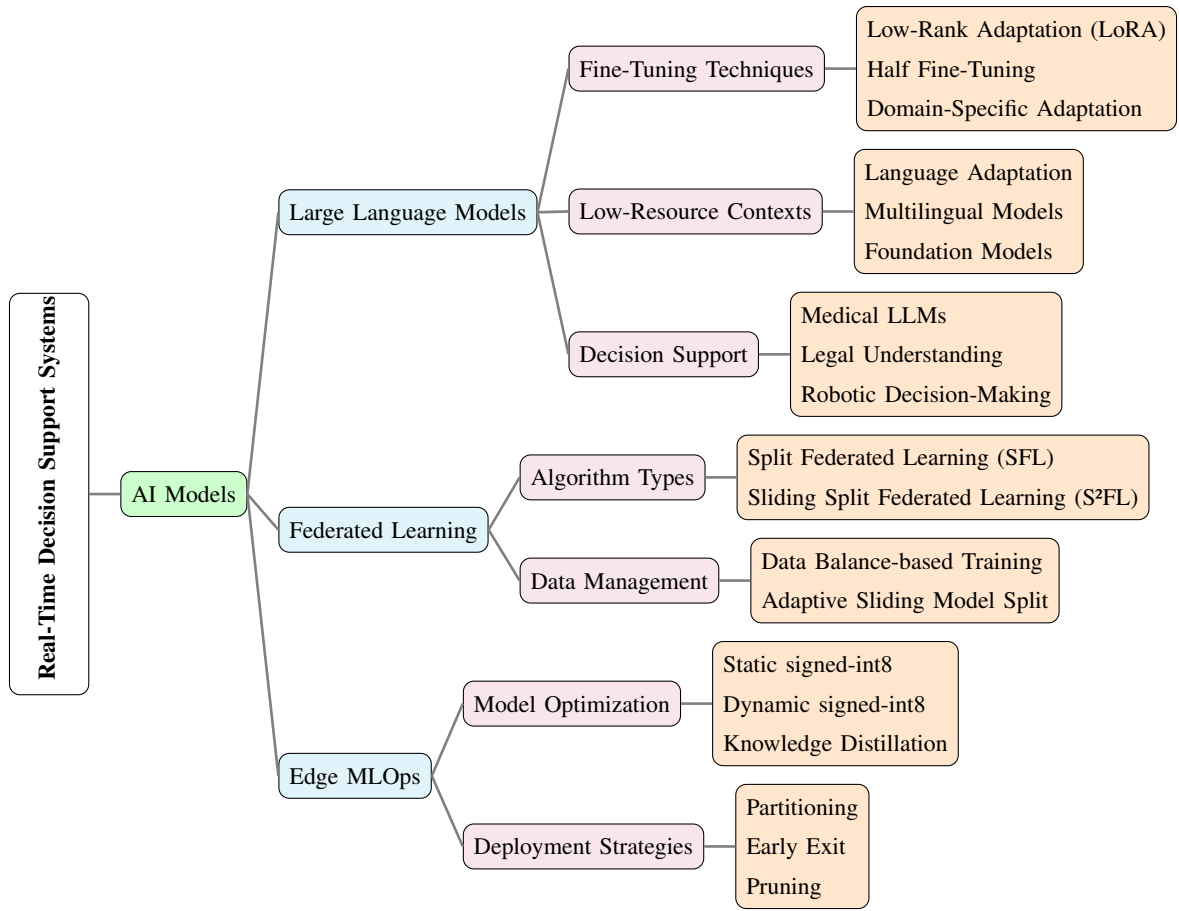

\section{Future Directions in Human-AI Collaboration with Edge Computing}

Future directions in human-AI collaboration and edge computing are poised to revolutionize real-time decision support systems, enabling more efficient, adaptive, and versatile applications. The analyzed papers contribute significantly to this advancement, with key findings and contributions in areas such as communication-efficient edge AI inference \cite{2004.13351}, edge general intelligence via large language models \cite{2410.18125}, and edge-cloud polarization and collaboration \cite{2111.06061}. The use of game-theoretic solvers \cite{2409.07460} and fast edge-based synchronizer \cite{2021.93825} also demonstrates the diversity of technical approaches being explored.
These developments highlight the importance of edge computing in enabling low-latency, real-time AI applications, as well as the need for effective collaboration between humans and AI systems.

The papers also highlight the challenges and limitations associated with human-AI collaboration and edge computing, including resource constraints \cite{2004.13351}, scalability and complexity issues \cite{2410.18125}, and the need for effective synchronization and coordination mechanisms \cite{2021.93825}. A common theme among the papers is the emphasis on achieving low latency and energy efficiency in edge AI inference and real-time AI applications. This is evident in the development of wireless distributed computing frameworks \cite{2004.13351} and cooperative transmission strategies for low-latency and energy-efficient AI services. Addressing these challenges will be essential for realizing the full potential of edge computing and AI collaboration in real-time decision support systems. Additionally, the development of more adaptive and versatile applications will require a deeper understanding of human-AI interaction patterns \cite{2310.19778} and the integration of emerging technologies, such as blockchain \cite{2409.07460}.

In conclusion, future directions in human-AI collaboration and edge computing are characterized by a strong emphasis on achieving low latency, energy efficiency, and effective collaboration between humans and AI systems. The development of communication-efficient edge AI inference, edge general intelligence via large language models, and edge-cloud polarization and collaboration will be crucial for advancing real-time decision support systems. Addressing the challenges and limitations associated with these developments, including resource constraints, scalability issues, and synchronization mechanisms, will be essential for realizing their full potential. As the field continues to evolve, we will likely see significant advancements in areas such as human-AI interaction patterns, blockchain-based security, and the integration of emerging technologies, ultimately leading to more efficient, adaptive, and versatile applications in a wide range of domains \cite{2111.06061}, \cite{2410.18125}.

\section{Conclusion}

The future of artificial intelligence (AI) and its applications in real-time decision support systems using low-latency AI models holds tremendous promise, as evident from analyzing the relevant studies. The technical details and methodologies employed in these studies are varied, ranging from machine learning techniques to systematic reviews and user studies. For example, the development of DISS relied on machine learning algorithms, while the systematic review of decision support systems in fisheries and aquaculture employed a comprehensive literature search and analysis methodology \cite{1611.08374}. User studies have also been conducted to test the efficacy of behavior descriptions in improving human-AI collaboration \cite{2301.06937}. Other notable studies have investigated the impact of AI-enhanced decision support on operator states and control rooms \cite{2402.13219}, as well as the determinants of LLM-assisted decision-making \cite{2402.17385}. Additionally, researchers have explored the use of multi-AI complex systems in humanitarian response \cite{2208.11282} and the development of uncertainty-aware resource management for real-time inference of language models \cite{2309.06619}. These studies demonstrate the diversity and complexity of research in this field, with many opportunities for further exploration and innovation.

However, there are also limitations and challenges associated with AI, such as the need for improved data quality and availability, the risk of suboptimal performance, and the importance of setting realistic goals and providing informed opinions \cite{2501.06086}, \cite{2209.03990}. In the context of real-time decision support systems, the development of AI models that can learn from data and adapt to changing environments is essential \cite{2110.05234}. The development of various sub-components, including data collection, data conditioning, algorithms, computing, robust AI, and human-machine teaming \cite{1905.03592}. The implications of these developments are far-reaching. For instance, the use of edge computing and IoT integration can enable real-time monitoring and decision making in healthcare \cite{1910.05316}, while also improving the efficiency and security of financial transactions \cite{2409.07460}. Moreover, the integration of human-AI collaboration and edge computing can facilitate more efficient navigation in constrained environments \cite{2201.07050} and enhance the overall decision-making process \cite{2403.05911}.. Furthermore, the consideration of ethical and societal implications of AI is crucial to ensure that the benefits of AI are equitably distributed and that the risks are mitigated \cite{2209.03990}, \cite{2501.07432}.

The analysis of papers also reveals contrasting viewpoints or approaches, such as the focus on technical details versus broader implications, and optimism versus caution \cite{1804.01396}, \cite{2501.06086}. Despite these differences, there is a consensus on the potential benefits of AI in real-time decision support systems, including improved accuracy, efficiency, and scalability \cite{1905.03592}, \cite{2209.03990}. The use of low-latency AI models can enable faster decision-making, which is critical in applications such as healthcare, finance, and transportation \cite{2006.03434}, \cite{2411.13562}.

In conclusion, the development of real-time decision support systems using low-latency AI models is a vibrant and rapidly evolving field, with many exciting developments and opportunities for further research. By exploring the key themes and connections between these studies, we can gain a deeper understanding of the complex relationships between human-AI collaboration, interpretability, and decision-making outcomes. Ultimately, this knowledge will enable us to create more effective and efficient decision support systems that can be applied in a wide range of domains, from healthcare to finance.




\bibliographystyle{apalike}
\bibliography{cite_dblp}

\end{document}